\documentclass{elsarticle}
\usepackage[utf8]{inputenc}
\usepackage[english]{babel}

\usepackage{times}
\usepackage{nohyperref}
\usepackage{url}
\usepackage{graphicx,color}
\usepackage{epsfig}
\usepackage[algoruled,linesnumbered]{algorithm2e}
\usepackage{algorithmic}
\usepackage{amsmath}
\usepackage{amssymb}
\usepackage{array}
\usepackage{subfigure}
\usepackage{verbatim}
\usepackage{flushend}
\usepackage{booktabs}
\usepackage{nomencl}
\usepackage{multirow}

\usepackage{savesym}
\savesymbol{pdfbookmark}
\usepackage{hyperref}
\restoresymbol{HR}{pdfbookmark}
\usepackage{float}
\makenomenclature  
 \newtheorem{theorem}{Theorem}[section]

 \newtheorem{definition}{Definition}[section]

\journal{Signal Processing}

\begin{document}
\begin{frontmatter}
\title{Hierarchical Symbolic Dynamic Filtering of Streaming Non--stationary Time Series Data}
\author[]{Adedotun Akintayo$\dag$}
\author{Soumik Sarkar$\dag$\corref{corres}}
\ead{soumiks@iastate.edu}
\cortext[corres]{Corresponding author. Tel: +1 515 357 4328}
\address{$\dag$Department of Mechanical Engineering, 2025 Black Engineering\\
Iowa State University, Ames, IA 50011, USA}

\begin{abstract}
This paper proposes a hierarchical feature extractor for non--stationary streaming time series based on the concept of switching observable Markov chain models. The slow time--scale non--stationary behaviors are considered to be a mixture of quasi--stationary fast time--scale segments that are exhibited by complex dynamical systems. The key idea is to model each unique stationary characteristics without \textit{a priori} knowledge (e.g., number of possible unique characteristics) at a lower logical level, and capture the transitions from one low--level model to another at a higher level. In this context, the concepts in the recently developed Symbolic Dynamic Filtering is extended, to build an online algorithm suited for handling quasi--stationary data at a lower level and a non-stationary behavior at a higher level without \textit{a priori} knowledge. A key observation made in this study is that the rate of change of data likelihood seems to be a better indicator of change in data characteristics compared to the traditional methods that mostly consider data likelihood for change detection. Thus, an adaptive Chinese Restaurant Process ($CRP$) distribution is formulated to accommodate the rate of change of data likelihood in the learning algorithm. The algorithm also aims to minimize model complexity while capturing data characteristics (likelihood). Efficacy demonstration and comparative evaluation of the proposed algorithm are performed using time series data simulated from systems that exhibit nonlinear dynamics. We discuss results that show that the proposed hierarchical symbolic dynamic filtering ($HSDF$) algorithm can identify underlying features with significantly high degree of accuracy, even under very noisy conditions. Based on our validation experiments, we demonstrate better performance of the proposed fast algorithm compared to the baseline Hierarchical Dirichlet Process--Hidden Markov Models (HDP--HMM).
The proposed algorithm's low computational complexity also makes it suitable for on--board, real time operations.
\end{abstract}
\begin{keyword}
Hierarchical Symbolic Dynamic Filtering; Deep Feature Extraction; adaptive Chinese Restaurant Process ; Stickiness; Likelihood change rate
\end{keyword}
\end{frontmatter}

\clearpage
\mbox{}
\printnomenclature

\section{Introduction}\label{sec:intro}
A challenging task in most statistical signal processing and machine learning applications is that of extracting informative features from spatial, temporal or spatiotemporal observations. For example, data-driven analysis of dynamical systems involve identification of salient features from the time series output data that may possess several intermixed modes or features. These problems are usually further complicated by the lack, or inadequacy, of ground truth labels to enable supervised learning from signals. Also, the absence of \textit{a priori} knowledge of how many unique characteristics are embedded in the data sets has been a critical challenge that has been explored in depth by the nonparametric modeling community over the past few years.

Among the different approaches proposed by the nonparametric modeling community, hierarchical feature extraction emerged as one of the most effective processes to solve the problem. The task of extracting hierarchical features also appears as a primary technical challenge in applications such as complex system modeling, robotics and image processing. In autonomous perception problems for instance, machines are required to learn to interact with human users and perform certain tasks by being aware of user information without being provided explicit commands. Such problems ~\cite{DRY11} require that the sensed information are mapped to the right contexts based on their explanatory features. In this context, one of the key innovations that emerged from the deep learning community is, scalable learning of hierarchical features as neural network parameters (i.e., weights and biases) with regularized backpropagation algorithms.
Complex feature extraction tasks from multidimensional data such as those in image, video and signal processing applications have been addressed deep learning approaches~\cite{HS06,BO11,LAS17,AKSS16, ANVMCAAGBS16,KDMBS15} with significant success. While these approaches have been efficient in the representation learning aspect, the challenges mostly arise from large computational burden due to the large parametric space of the hierarchical models.

Specifically, in the domain of time series data, deep recurrent neural networks (RNN)~\cite{ZJC15} and long short term memory (LSTM)~\cite{FGJSFC00} networks have shown significant promise. However, they require significantly large amount of labeled data for supervised training process. On the unsupervised side, Hierarchical Dirichlet Process--Hidden Markov Model (HDP-HMM) technique~\cite{FSJW11} has emerged as one of the primary tools for solving classical problems such as the speaker diarization problem~\cite{TR06}. This problem requires the identification of features that represent `who spoke what ?' and  `when ?' from speech time series data in an unsupervised manner. The problem can be formulated as a standard hierarchical feature extraction problem where the number of unique speakers may be identified from the signal and represented by the number of features. Cognitive processes in humans also show that ideas are generated in an adaptive and hierarchical manner. This conjecture has encouraged a lot of interest and improved modeling of human learning and reasoning processes using probabilistic programming concepts~\cite{TKGG11}. However, due to the involvement of latent space evaluation in techniques such as HDP-HMM, they tend to be rather slow and thus difficult to use in on-board, real time applications.

For fast time series feature extraction, a method called Symbolic Dynamic Filtering ($SDF$) was proposed recently that can be categorized as an observable Markov chain model. The technique was shown ~\cite{SSM13} to be efficient for learning dynamical systems and has been found to perform better in terms of anomaly detection under noisy environment compared to techniques such as (shallow) Artificial Neural Networks ($ANN$), Principal Component Analysis ($PCA$) and Bayesian filtering techniques~\cite{RRSY09}. Successful applications of $SDF$ includes a variety of complex systems such as nuclear power plants~\cite{JGSR11}, coal--gasification systems~\cite{CSGR08}, ship--board auxiliary systems~\cite{SNMAS13} and gas turbine engines~\cite{SYGRM08,GRSY08}. More recently, the framework was extended to model multivariate interactions via spatio--temporal pattern network formulation~\cite{CSZS16}, and have been applied to characterize wind turbine interactions~\cite{ZS15}, bridge monitoring using dense sensor networks~\cite{LGLPS17} as well as complex cyber--physical systems~\cite{SZASA16}. In the SDF algorithm, the absence of the space of latent variables in Markov chain modeling ensures low memory consumption and enhanced computational efficiency~\cite{SSR13}. Therefore, it can be used for on-board, real time learning and adaptation which may be an issue for deep learning and nonparametric techniques. Because $SDF$ approximates a symbolic time series as a Markov chain of certain order; the modeled time series is implicitly assumed to be statistically stationary~\cite{R04} at a slow time epoch. However, this assumption is a prohibitive one while dealing with non--stationary characteristics of data even at a slow time--scale. Note, by non--stationary characteristics at a slow time--scale, we mean that the time series contains multiple quasi--stationary characteristics at a fast time--scale and hence becomes non--stationary when viewed at a slower time--scale.

To solve this problem, a hierarchical SDF (HSDF) framework~\cite{AS15} can be formulated that can model different quasi--stationary characteristics using different SDF models (in the form of Probabilistic Finite State Automata ($PFSA$)). The entire time series can then be expressed as a higher level $PFSA$ whose states are the automata obtained for different unique characteristics. Regarding hierarchical modeling with SDF, a Multi--scale Symbolic Time Series Analysis ($MSTSA$)~\cite{SDR15} approach has been proposed recently for characterizing seismic activities monitored by Unattended Ground Sensor ($UGS$) in an online manner. The main difference between the method presented in that paper and the one proposed here is the lack of labels or supervision (i.e., without knowing how many unique characteristics or classes are present in data as well as no knowledge of the number and period of transition between unique classes are present) that is handled in the present case.

Note that while the proposed modeling architecture is able to model nonlinear systems, they share similar goal with that in~\cite{FSJW09a} for learning the switching linear dynamical systems ($SLDS$). To this end, validation and comparison of efficacy of the proposed approach are done on time series generated from nonlinear dynamical systems based on the chaotic Duffing and Van der Pol equations~\cite{RMSR09}.

The main contributions of this paper are:
\begin{itemize}
\item Development of a novel Hierarchical Symbolic Dynamic Filtering ($HSDF$) algorithm that can model a non--stationary time series composed of several quasi--stationary behaviors where neither the behaviors nor the number of unique characteristics are known.
\item Demonstration of the effects of various concepts such as adaptive Chinese Restaurant Process (CRP), likelihood change rate and Stickiness adjustment on the $HSDF$ performance.
\item Development of an off--line PFSA model revision strategy to improve HSDF performance.
\item Testing and validation of the proposed algorithm, as well as performance comparison with $SLDS$ generated by the sticky HDP--HMM approach in order to show that the proposed algorithm is a fast and computationally efficient method with potential usefulness for application in real--life feature extraction.
\end{itemize} 
Beyond this section, the paper is presented using five more sections as follows. A brief review of the $SDF$ framework as well as other major statistical components are provided in Section \ref{sec:background}. Formulation and implementation of the online $HSDF$ algorithm are presented in Section~\ref{sec:Methodology}. Some analytical results related to data likelihood improvement using the proposed algorithm is provided in Section~\ref{sec:Proof}. Section~\ref{sec:validation} provides validation results and discussions of experiments on simulated nonlinear dynamical systems. Finally, Section~\ref{sec:con} summarizes and concludes the paper along with recommendations of future work.

\section{Background and Motivation} \label{sec:background}
Time series signals obtained from dynamical systems can be decomposed into multiple (two in this case) time--scales as described in~\cite{AS15}. The implication of such decomposition is that a quasi--stationary signal that is acquired at some fast time--scale can be effectively modeled by a Probabilistic Finite State Automaton ($PFSA$) using symbolic dynamic filtering (SDF)~\cite{R04}. 

\begin{figure}
\centering
\includegraphics[width=0.75\textwidth]{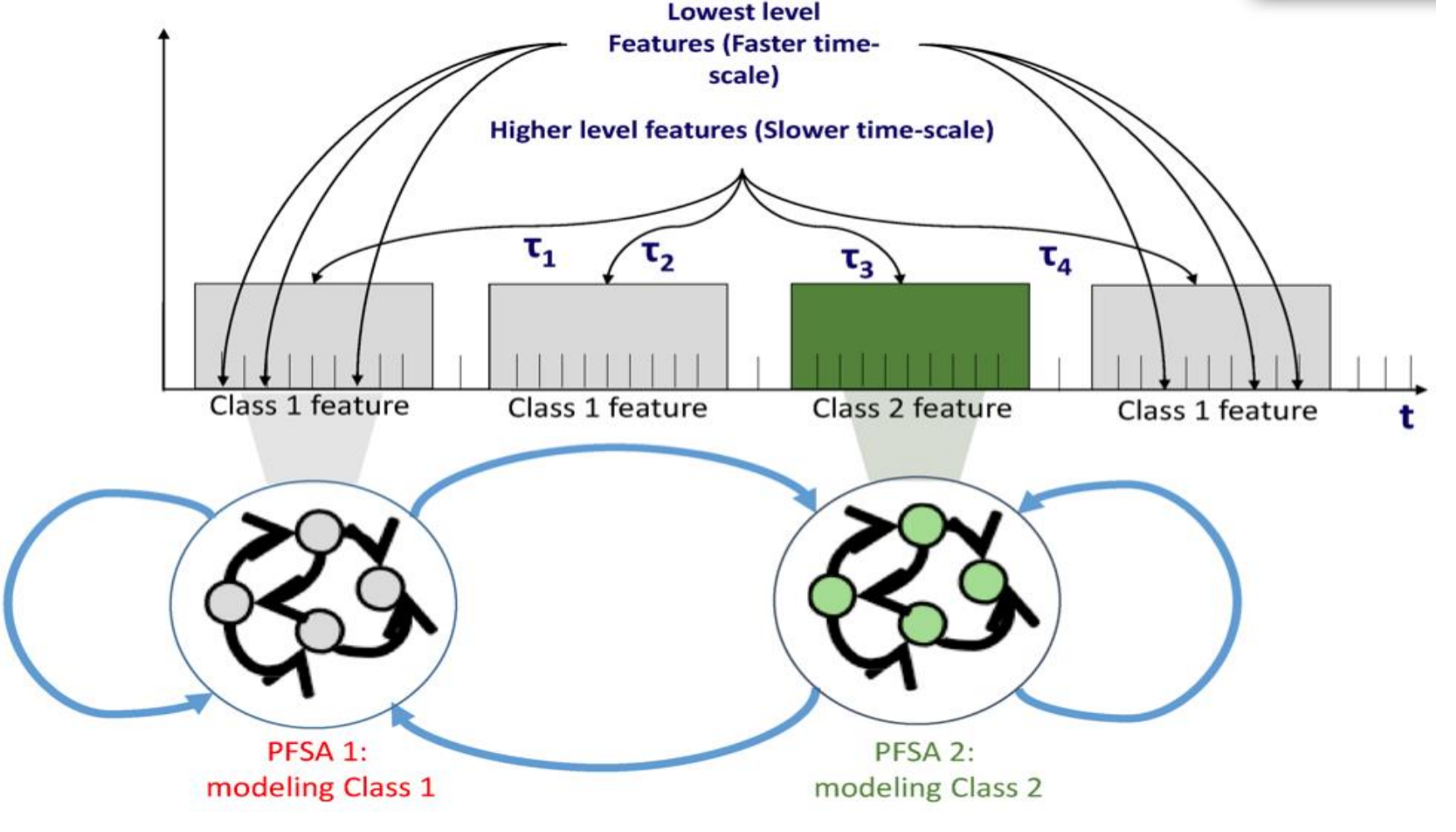}  
\caption{\textit{Schematics of Hierarchical PFSA based Feature Extraction.}}
\label{Fig:deepsdf}
\end{figure}

However, at a slower time--scale, the time series can be non--stationary in nature and the standard SDF scheme is not suited to handle it. With this motivation we introduce the concept of hierarchical symbolic dynamic filtering (HSDF) using an illustrative example described in Figure \ref{Fig:deepsdf}. We begin with slow time epoch $\tau_1$ that is represented by the PFSA 1 model. Upon learning PFSA 1 from slow time epoch $\tau_1$ (using standard SDF), we move on to the second slow time epoch $\tau_2$. In this example, $\tau_2$ belongs to the same quasi--stationary characteristics as in $\tau_1$. Hence, the challenge is to identify the similarity and classify $\tau_2$ as a member of class 1 represented by PFSA 1. Note, this class retention mechanism can be perceived as a self-transition at a higher logical level where PFSA 1 is considered as a state of the underlying system. After this, let us assume that slow time epoch $\tau_3$ belongs to a new, unforeseen quasi--stationary characteristics. In this case, we need to identify the change in characteristics from the streaming data and create a new PFSA to represent the characteristics. At the higher logical layer, this becomes a state transition from state/class 1 (represented by PFSA 1) to state/class 2 (represented by PFSA 2). Therefore, we start learning an upper-level PFSA to model class retentions and transitions for the system. Time series from the underlying system can thus be described in a bottom--up hierarchical manner where the goal is to discover multiple $PFSA$s at the lowest level which in turn become the states of an upper-level $PFSA$. Notionally, this is similar to the concept of deep learning where at a upper layer (i.e., beyond the first hidden layer), 'features of features' are learnt from data for an efficient representation.
A thorough review of the $SDF$ framework can be found in ~\cite{R04,GRSY08}. However, for the purpose of completeness, a brief description of $SDF$ and the framework's other main constituents (namely, CRP and stickiness factor) are presented in Subsection~\ref{sec:SDF}.

\subsection{Symbolic Dynamic Filtering Formulation}\label{sec:SDF}
Dynamical systems generate time series data which lie in the space of continuous or discrete signals. In the symbolic dynamic filtering literature~\cite{R04}, quantization of the continuous (in this case streaming--type) signals (or in some cases already discrete signals) into symbol sequences is a major first step in the $SDF$ formulation. There are many ways of quantization (or partitioning as called in the Symbolic Dynamics literature) reported in the literature~\cite{SS16,SSS13,SMJ12} depending on different objective functions. However, the focus of this paper is to model symbol sequences (using $PFSA$ models) obtained after such quantization.

Given a suitably defined nonempty, finite set of symbols called alphabet $\Xi$, and nonempty, finite set of states $\Theta$, we define a $PFSA$, $\mathcal{P}$ as a 4--tuple, such that $\mathcal{P} \triangleq(\Theta,\Xi,\delta,\Omega)$. Nonlinearities in the time series are represented by a specific type of $PFSA$ called the D--Markov machines~\cite{MR14} where past depth D of symbols are considered for modeling the states as given by $|\Theta|\leq|\Xi|^D$~\cite{R04}. The mapping $\delta: \Theta \times \Xi \rightarrow \Theta$ denotes a function that maps the transitions from a current state to a future state (or self transition) given the alphabet. Also, we consider a morph function $\pi: \Theta \times \Xi \rightarrow [0,1]$ that satisfies the condition $\sum_{\xi \in\Xi}\pi(\theta,\xi)=1$. Based on the morph function, we define the non-negative $(|\Theta|\times |\Xi|)$ state transition matrix $\Omega$ as: $\Omega_{ij} \triangleq \pi(\theta_i,\xi_j),\forall \theta_i\in \Theta$ and $\forall \xi_j\in\Xi$. Online learning of an $SDF$ model involves identifying this matrix.
\nomenclature{$\Xi$}{Finite alphabet}%
\nomenclature{$\Theta$}{Finite set of states}%
\nomenclature{$\delta$}{State transition function}%
\nomenclature{$\Omega$}{State transition matrix}%
\nomenclature{$\theta$}{Symbolic state}%
\nomenclature{$\xi$}{Symbol}%
\nomenclature{$D$}{Depth (symbolic states are strings of $D$ symbols)}%

Note that initial state $\theta_0 \in \Theta$ of the quasi--stationary data represented by the $\mathcal{P} \triangleq(\Theta,\Xi,\delta,\Omega)$ have no influence on the state transition. However, a simple frequency count of the occurrence of symbols in the training string sequence at depth $D$, followed by normalization is used to derive the low dimensional encoding matrix $\Omega$. With that knowledge, new testing symbol strings that also follow the same quantization can be evaluated for similarity or difference with a $PFSA$ represented by $\Omega$.
\subsection{The CRP distribution and Stickiness adjustment}\label{sec:CRP}
This subsection briefly describes a couple of basic statistical concepts used in the proposed formulation, namely the Chinese Restaurant Process ($CRP$) and data likelihood adjustment using the stickiness factor. Recently, these ideas have been extensively used in nonparametric modeling techniques and therefore details can be found in the related literature~\cite{W07,FSJW11}.

\begin{figure}
\centering
\includegraphics[width=\textwidth]{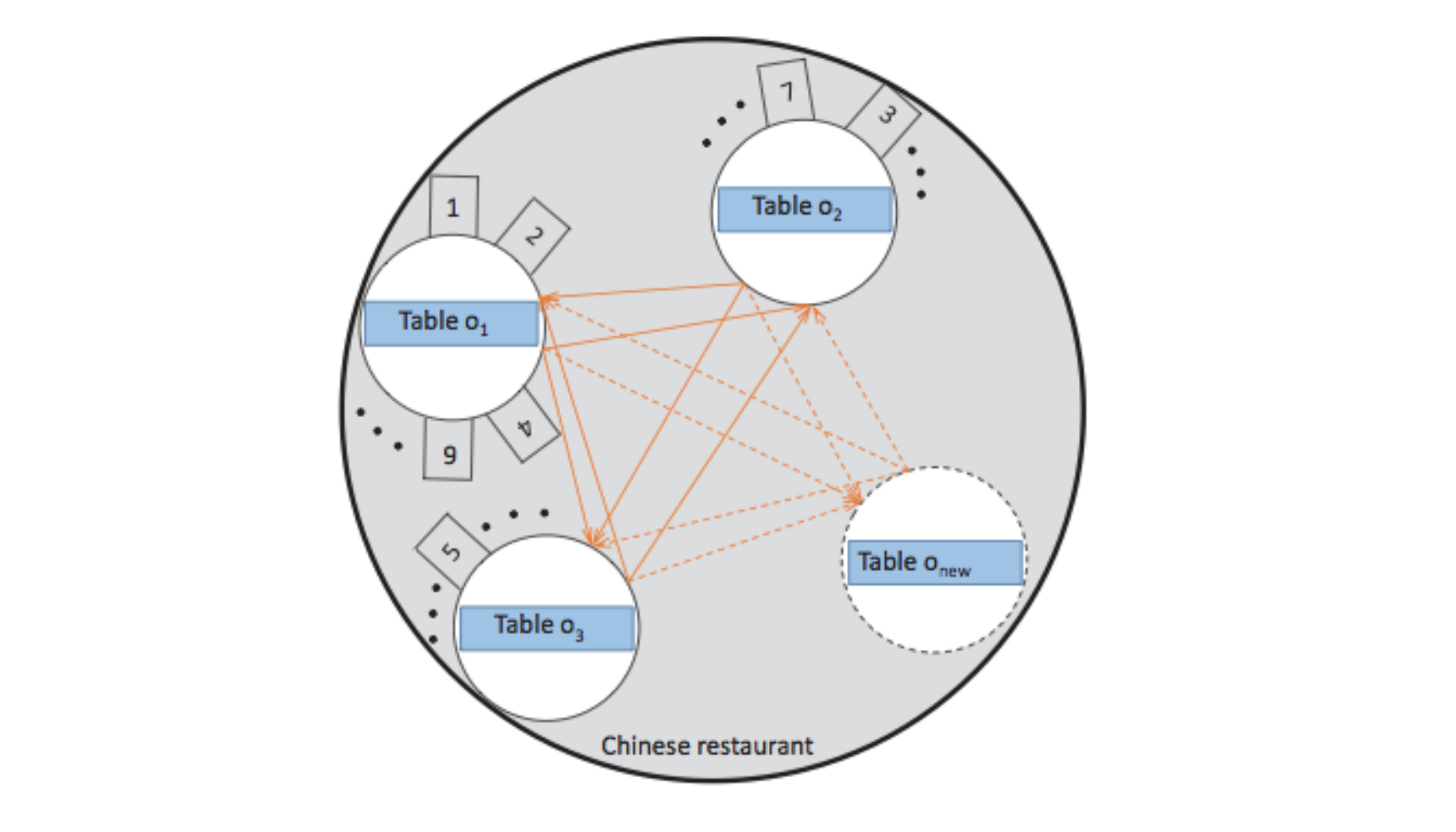}  
\caption{\textit{Illustration of the Chinese Restaurant Process with numbers: 1, 2, 3, $\cdots$ indicating customers' order of arrival.}}
\label{Fig:CRP} 
\end{figure}

$CRP$ represents a discrete sequence over partitions that is suitable for modeling infinite mixtures, hence often used for modeling clusters in Bayesian frameworks. $CRP$ shares some similarities with the stick breaking and the Dirichlet Process, but with some subtle differences in how the processes evolve~\cite{W07}. The crux of the $CRP$ distribution is to model the tendencies of newly arriving customers to a fictional Chinese restaurant to either sit in an existing table $\in O$ or in a new table, $o_{new}$~\cite{A85} with less restriction on number of tables or customers at a table (as illustrated in Figure \ref{Fig:CRP}). Therefore, $CRP$ is a suitable candidate for nonparametric modeling. The CRP distribution can be mathematically described as follows:
\begin{equation}
Pr_{\epsilon}(o\in O) = \frac{\mathbb{C}(o)}{[\sum_{x\in O} \mathbb{C}(x)]+\epsilon}
\label{equ:eps}
\end{equation}
\begin{equation}
Pr_{\epsilon}(o_{new}) = \frac{\epsilon}{[\sum_{x\in O} \mathbb{C}(x)]+\epsilon}
\label{equ:epsilon}
\end{equation}

where $\mathbb{C}(\cdot)$ signifies a concentration function and $\epsilon$ is called the $CRP$ parameter. Naturally, this paper uses a $CRP$ process to determine whether a new $PFSA$ model is required to model a newly arriving slow time epoch or an existing PFSA would suffice.

While induction of $CRP$ can help in deciding the need for a new $PFSA$ model, noise and spurious disturbance present in real data can drive the decision system to instability. That is, many unnecessary new $PFSA$ models may get generated and the decision may then fluctuate among different $PFSA$ models that are close to each other, with closeness based on an appropriate metric. Similar situation arises in other unsupervised techniques as well such as the HDP--HMM. Assuming inference could be made after the arrival of several slow time epochs, an off--line revision that is described in Subsection\ref{sec:PFSA} will be effective at merging such spurious classes. However, in most real--life application, decisions have to be taken in an online manner, i.e., soon after the arrival of the new slow time epochs. A stickiness factor described by~\cite{FSJW11, AS15} was found to be effective for reducing such fluctuations. The extra information incorporated by stickiness is to lean slightly towards the class occupied by the most recently assigned (before the current epoch) slow time epoch. The ideas discussed in this section are described mathematically in Section \ref{sec:Methodology}.
\nomenclature{$B(\cdot)$}{Normalization constant for likelihood estimation}%
\nomenclature{$\epsilon$}{$CRP$ hyperparameter}%

\section{Proposed Hierarchical $SDF$ Methodology}\label{sec:Methodology}
In this section, we describe the proposed hierarchical symbolic dynamic filtering ($HSDF$) framework along with the learning scheme using streaming non--stationary time series data. 
In general, two technical challenges are involved in such a problem involving streaming data. The first is deals with the reliability (or accuracy) of the inference for decision--making, while the second is how quickly the inference can be made. 
In this paper, we aim to present a comprehensive approach tackling both challenges for robust decision--making. Most real--life dynamical systems especially those with safety, security, reliability or dependability concerns~\cite{SZASA16, SYGRM08,GRSY08, JGSR11, SSM13} require such online inference capability for decision--making.\\
We begin with a mathematical derivation of the data likelihood of a newly arriving slow time epoch given a few existing classes (PFSA models) in Subsection \ref{sec:pro}. Formulation of an adaptive $CRP$ process and stickiness factor for assigning slow time epochs to existing and new classes are provided in Subsection \ref{sec:assignment}. Finally, the outline of an off--line algorithm for periodic revision (over a few slow time epochs) of the space of PFSA models learnt by the online process is presented in Subsection \ref{sec:PFSA}.
\subsection{Data Likelihood estimation} \label{sec:pro}
Let us assume that $K$ classes representing $K$ unique quasi--stationary behavior occurred in the past data epochs have already been identified. Let the distinct set of classes be $\{C^i: i = 1,2,\cdots,K \}$, over the same sets of symbol $\Xi$ and state $\Theta$, and each class $C^i$ is modeled by a PFSA $\mathcal{P}^i=(\Theta^i,\Xi,\delta^i,\Omega^i)$. Also, let symbol strings belonging to the class be $S^i\triangleq s^i_1 s^i_2\ldots$. An appropriate depth $D$ is selected for the $D$--Markov machine from which the morph (probability) matrix $\Omega^i$ has been derived. Each row of the $\Omega$ is normalized in order to perform inference on a new slow time epoch.\\
Let the $m^{th}$ row of $\Omega^i$ be denoted as $\Omega^i_m$ and the $n^{th}$ element of the $m^{th}$ row as $\Omega^i_{mn}\geq 0$ and $\sum_{n=1}^{|\Xi|} \Omega^i_{mn}= 1$. The \emph{a priori} probability density function $f_{\Omega^i_m|S^i}$ of the random row-vector $\Omega^i_m$ that is conditioned on a symbol string $S^i$ can be modeled by the Dirichlet distribution~\cite{TSF73,JS94} as,
\begin{align}\label{equ:fu}
 f_{\Omega^i_m|S^i}(\boldsymbol{\omega}_m^i|S^i)=\frac{1}{B(\boldsymbol{\alpha}^i_m)}\prod_{n=1}^{|\Xi|} (\omega^i_{mn})^{\alpha^j_{mn}-1}
\end{align}  
where, each column of the $\Omega^i$ is represented by $\boldsymbol{\omega}^i_m$ as, 
$$\boldsymbol{\omega}_m^i =\begin{bmatrix}
\omega^i_{m1} & \omega^i_{m2} & \ldots & \omega^i_{m|\Xi|}
\end{bmatrix}   
$$
and the constant for normalization is
\begin{align} \label{equ:const}
 B( \boldsymbol{\alpha}^i_m )\triangleq\frac{\prod_{n=1}^{|\Xi|} \Gamma(\alpha^i_{mn})}{\Gamma (\sum_{n=1}^{|\Xi|} \alpha^i_{mn})}
\end{align}  
where, $\Gamma(\bullet)$ denotes the gamma function, and
$\boldsymbol{\alpha}^i_m=\begin{bmatrix}
\boldsymbol{\alpha}^i_{m1} & \boldsymbol{\alpha}^i_{m2} & \ldots & \boldsymbol{\alpha}^i_{m|\Xi|}
\end{bmatrix}
$ with
\begin{align}\label{equ:alpha}
\alpha^i_{mn}=N^i_{mn}+1
\end{align}  
At a state $\theta_m$, the number of times the symbol $\xi_n$ arises in the transition to a new state is modeled by $N^i_{mn}$ as, 
\begin{align}
 N^i_{mn} \triangleq \big|\{(s_k^i,v^i_k):s_k^i=\xi_n,v^i_k=\theta_m\}\big|
\end{align}  
where $s_k^i$ is the $k^{th}$ symbol in $S^i$ and $v_k^i$ is the $k^{th}$ state as obtained from the symbol sequence $S^i$. Note that a state is defined as a sequence of past D symbols. $N^i_m \triangleq \sum_{n=1}^{|\Xi|}N^i_{mn}$ computes the number that state $\theta_m$ occur in the state sequence. From Equation \ref{equ:const} and Equation \ref{equ:alpha}, it follows that
\begin{align} \label{equ:normalConstant}
 B( \boldsymbol{\alpha}^i_m )=\frac{\prod_{n=1}^{|\Xi|} \Gamma(N^i_{mn}+1)}{\Gamma (\sum_{n=1}^{|\Xi|} N^i_{mn}+|\Xi|)}
 =\frac{\prod_{n=1}^{|\Xi|}(N^i_{mn})!}{\left( N^i_m + |\Xi|-1\right)!}
\end{align}  
where the standard definition, $\Gamma(n)=(n-1)! \ \ \forall n \in \mathbb{N}$ has been used.\\
Markov property of the $PFSA \ \mathcal{P}^i$, ensures that the $(1\times|\Xi|)$--dimension row vectors of $\Omega^i$ are statistically independent of each other, $\forall m=1,\ldots |\Theta|$. Equation \ref{equ:const} and Equation \ref{equ:normalConstant} thus lead to conditioning of the \emph{a priori} joint density, $f_{\Omega^i|S^i}$ of the probability morph matrix, $\Omega^i$  on the symbol string, $S^i$ as,
\begin{align} \label{equ:fi}
f_{\Omega^i|S^i}(\boldsymbol{\omega}^i|S^i)
&=\prod_{m=1}^{|\Theta|} f_{\Omega^i_m|S^i}\left(\boldsymbol{\omega}_m^i|S^i \right)
\notag \\ & = \prod_{m=1}^{|\Theta|}
\left( N^i_m + |\Xi|-1\right)!
 \prod_{n=1}^{|\Xi|} \frac{(\boldsymbol{\omega}_m^i)^{N^i_{mn}}}{(N^i_{mn})!}
\end{align}  
where, $\boldsymbol{\omega}^i= \Big[(\boldsymbol{\omega}_1^i)^T \ (\boldsymbol{\omega}_2^i)^T \ \cdots \ (\boldsymbol{\omega}_{|\Theta|}^i)^T\Big] \in [0,1]^{|\Theta|\times|\Xi|}$, and $T$ is a transpose operator.\\
At this point, if a new slow time test epoch is represented by $\widetilde{S}$, its probability of belonging to a certain $PFSA$ model $(\Theta,\Xi,\delta,\Omega^i)$, given the morph matrix $\Omega^i$ derived from the training symbol sequence $S^i$ can be represented as a product of multinomial distributions~\cite{W63} as,
\begin{align}
& \quad \Pr\left(\widetilde{S}|\Theta,\delta, \Omega^i \right) \notag
\\ & = \prod_{m=1}^{|\Theta|} (\widetilde{N}_m)!\prod_{n=1}^{|\Xi|}\frac{\left(\Omega^i_{mn}\right)^{\widetilde{N}_{mn}}}{(\widetilde{N}_{mn})!} \label{equ:multinomial}
\\ & \triangleq \Pr\left(\widetilde{S}|\Omega^i \right) \ \ \textrm{as} \ \Theta \ \textrm{and} \ \delta \ \textrm{are kept invariant}
\label{equ:notation}
\end{align} 
where, given a state $\theta_m$, the number of times the symbol $\xi_n$ present in the testing string $\widetilde{S}$ occurs during transition to a new state is modeled by $\widetilde{N}_{mn}$ as,
\begin{align}
  \widetilde{N}_{mn} \triangleq \big|\{(\tilde{s}_k,\tilde{v}_k):\tilde{s}_k=\xi_n,\tilde{v}_k=\theta_m\}\big| \vspace{-12pt}
 \label{equ:countm}
\end{align} 
where again, the $k^{th}$ symbol in the observed string $\widetilde{S}$ is $\tilde{s}_k$, and the $k^{th}$ state derived from $\widetilde{S}$ is denoted by $\tilde{v}_k$.\\
Now, Equation \ref{equ:fi} and Equation \ref{equ:multinomial} can be combined to obtain the probability of a symbol string $\widetilde{S}$ belonging to a class characterized by already observed symbol string $S^j$. With the derivation presented by~\cite{SMSR00}, the following conditional distribution was obtained,
\begin{align}
\mu(\widetilde{S}| S^i) &=
 \prod_{m=1}^{|\Theta|} \frac{ (\widetilde{N}_m)!\left( N^i_m + |\Xi|-1\right)! }{ \left(\widetilde{N}_{m}+N^i_m + |\Xi|-1\right)!}
 \nonumber \\
 & \quad \times  \prod_{n=1}^{|\Xi|} \frac{ (\widetilde{N}_{mn}+N^i_{mn})! }{ (\widetilde{N}_{mn})!(N^i_{mn})! }\label{equ:ssq}
\end{align} 
where $\widetilde{N}_m \triangleq \sum_{n=1}^{|\Xi|}\widetilde{N}_{mn}$.
In practice, Stirling's approximation for the logarithm of a factorial $\log(n!)\approx n\log(n)-n$~\cite{P96} is mostly easier to compute, especially when either (or both) of $N^i$ and $\widetilde{N}$ consist of statistically large enough sample points (but still not be enough to directly estimate a $\Pi$ matrix at the testing phase). At this point, the likelihood probability, $\Pr(\widetilde{S}| S^i)$ may be easily found  by normalizing the conditional factors in Equation \ref{equ:ssq}.
\nomenclature{$K$}{Number of existing classes}%
\nomenclature{$S$}{The symbol sequence}%
\nomenclature{$\textbf{C}$}{Set of classes}%
\nomenclature{$\mu$}{Conditional data likelihood}%
\nomenclature{$\textbf{b}$}{$CRP$ adaptivity parameter}%
\nomenclature{$\textbf{A}(\cdot, \cdot)$}{Average rate of change of likelihood}%
\nomenclature{$\kappa$}{Stickiness factor}%
\nomenclature{$\nu$}{Threshold for likelihood rate change}%
\nomenclature{$\tau$}{Slow time epoch}%
\nomenclature{$\Delta$}{Memory parameter for likelihood rate estimation}%
\subsection{Assignment of a Slow time--scale behavior to an Existing or New class} \label{sec:assignment}
The objective of the inference process is to compute the probability of assigning a slow time epoch $\tau_j$ to a class $C^i \in \textbf{C}$ (where $\textbf{C} = \{C^i \ \forall \ i = 1, \cdots, K\}$ is the set of existing classes) or a newly created class $C^{K+1}$.
Let the symbol sequence for the current slow time epoch be $\widetilde{S}_{\tau_j}$.
Then the likelihood for class $C^i$ for current epoch $\tau_j$ is given by $\mu(\widetilde{S}_{\tau_j}|S^i)$ as described in the previous subsection.
The posterior probability for class selection can be denoted by $\Pr(C^i, S^i|\widetilde{S}_{\tau_j})$ which is equivalent to $\Pr(C^i|\widetilde{S}_{\tau_j})$ in this case since all existing classes are completely characterized by symbol sequences $S^i \ \forall i$.
With this setup, we obtain the following:
\begin{equation}
\Pr(C^i| \widetilde{S}_{\tau_j}) \propto \mu(\widetilde{S}_{\tau_j}|S^i) \ \ \forall i = 1, \cdots, K
\label{equ:prob1}
\end{equation}  
We use the Chinese Restaurant Process ($CRP$) to introduce the likelihood of a new class $C^{K+1}$ with a $CRP$ hyperparameter $\gamma_j$ as follows (note that the hyperparameter is specific to the test epoch $\tau_j$).
\begin{equation}
   \mu_{\gamma_j}(C^{K+1}|\widetilde{S}_{\tau_j}) = \gamma_j \sum_{i = 1}^{K} \mu(\widetilde{S}_{\tau_j}|S^i)
    \Rightarrow  \sum_{i = 1}^{K} \mu_{\gamma_j}(C^i|\widetilde{S}_{\tau_j}) = (1-\gamma_j) \sum_{i = 1}^{K} \mu(\widetilde{S}_{\tau_j}|S^i)
\label{equ:assign2}
\end{equation}
$CRP$ hyperparameter $\gamma_j$ (that was described in Equation \ref{equ:epsilon}) is given by the following expression.
\begin{equation}
 \gamma_j = \frac{\epsilon}{\left[\sum_{i = 1}^{K} \mu(\widetilde{S}_{\tau_j}|S^i)\right] + \textbf{b}\epsilon}
 \label{equ:epsilon2}
\end{equation}
where, $\epsilon \geq 0$ is a real valued parameter and $\mu(\widetilde{S}_{\tau_j}|S^i)$ is treated as the concentration or strength function found in Equation \ref{equ:eps} and Equation \ref{equ:epsilon} for the $CRP$ formulation. However, instead of the classical formulation \cite{AS15}, we introduce a new scalar multiplier $\textbf{b}$ that modifies the likelihood of creating a new class. The choice of $\textbf{b}$ will depend on a parameter $\textbf{A}(\Delta,i)$ that captures the rate of change of data likelihood as follows:
\begin{equation}
\textbf{A}(\Delta,i) = \frac{1}{\Delta}\sum_{p = 1}^{\Delta} \left[\mu(\widetilde{S}_{\tau_{j-p}}|S^i) - \mu(\widetilde{S}_{\tau_j}|S^i) \right]
\label{equ:prob2}
\end{equation}  
where $\Delta$ is a memory parameter that accommodates likelihoods of past epochs. It is evident from the expression of $\textbf{A}(\Delta,i)$ that it is essentially an expected reduction of likelihood of exisiting classes at the current epoch $\tau_j$. A high value of $\Delta$ reduces the noise in estimation, which can also reduce the senstivity to class changes. While $\textbf{A}(\Delta,i)$ can be incorporated in various ways to compute the $CRP$ parameter, it is accommodated in a discrete manner in the present formulation. Note that the condition with high values of $\textbf{A}(\Delta,i)\ \forall i$ suggests a significant drop in likelihood of all existing classes which increases the possibility of a new class generation. Therefore, a positive threshold, $\nu$ is chosen such that when $\textbf{A}(\Delta,i) > \nu \ \forall i$, we use $\textbf{b} = 1$ (i.e., classical formulation). Otherwise, we reduce the possibility of new class generation by taking $\textbf{b} = 2$ for the adaptive $CRP$ formulation.\\
At this point, we introduce the notion of `stickiness' in our proposed algorithm which is based on the fact that a real--life system usually may not fluctuate its operating point or internal parametric condition at each slow time epoch. In the present context, this means that if a slow time epoch $\tau_{j-1}$ belongs to a class, $C^k \in \textbf{C}$, then there will be a high likelihood for new streaming data at epoch $\tau_j$ to belong to $C^k$ as well. This notion is incorporated into the formulation by introducing a positive bias towards the last seen class $C^k$ as follows:
\begin{equation}
\mu_{\gamma_j}(C^{k}|\widetilde{S}_{\tau_j}) =  \max\left\{\frac{\kappa}{1 - \kappa} \sum_{i=1}^{K+1} \mu_{\gamma_j}(C^{i}|\widetilde{S}_{\tau_j}), \mu_{\gamma_j}(C^{k}|\widetilde{S}_{\tau_j})\right\}
\label{equ:dec}
\end{equation}
where $0 < \kappa < 1$ is the stickiness factor. Note, the rationale behind this adjustment is to ensure a certain minimum likelihood for the last seen class $C^k$ and in this context, the proposed formulation ensures that
\begin{equation}
\frac{\mu_{\gamma_j}(C^{k}|\widetilde{S}_{\tau_j})}{\sum_{i=1}^{K+1} \mu_{\gamma_j}(C^{i}|\widetilde{S}_{\tau_j})} \geq \kappa
\end{equation}
This can be proved by considering the extreme case when $\mu_{\gamma}(C^{k}|\widetilde{S}_{\tau_j}) = 0$, before applying the stickiness factor. Numerically, the `stickiness' adjustment significantly reduces the `hunting behavior' in class identification and creation process which will be demonstrated via numerical simulation results in the next section.\\
Finally, the $\mu_{\gamma_j}(C^{i}|\widetilde{S}_{\tau_j})$ factors are normalized to obtain the posterior probabilities $\Pr(C^{i}|\widetilde{S}_{\tau_j})$ for each class as follows:
\begin{equation}
\Pr(C^{i}|\widetilde{S}_{\tau_j}) =  \frac{\mu_{\gamma_j}(C^{i}|\widetilde{S}_{\tau_j})}{\sum \mu_{\gamma_j}(C^{i}|\widetilde{S}_{\tau_j})}
\label{equ:norm}
\end{equation}
We generate a random sample from this distribution to take a decision of class identification and generation at the testing epoch $\tau_j$.\\
The online algorithm for class assignment is summarized below. Note, we assume that partitioning and state construction are already performed before we begin the following algorithm. Hence, the alphabet $\Xi$ and state set $\Theta$ and the corresponding indices $n$ and $m$ are already defined.\\
\begin{algorithm}[H] \label{algo:1}
 \textbf{Input Parameters}: $CRP$ parameter $\epsilon$, memory parameter $\Delta$,\\
 likelihood rate threshold $\nu$ and stickiness parameter $\kappa$\\
  \textbf{Data Input}: Slow time epochs: $\tau_1, \tau_2, \cdots$ of symbolized string segments $\widetilde{S}_{\tau_l}$ \\
  \textbf{Initialization}: $\textbf{C} = \{C^1\}$ and Compute $N^1_{mn}$ using $\widetilde{S}_{\tau_1}$\\
  \ForAll{$\tau_2, \tau_3, \cdots$}{
   Compute $\widetilde{N}_{mn}$ using $\widetilde{S}_{\tau_j}$ \\
   \eIf{ $j < \Delta + 1$ }{
    Compute $\gamma_j$ using Equation \ref{equ:epsilon2} with $\textbf{b} = 2$ \\
   }{
   Compute $\textbf{A}(\Delta,i)$ for all existing classes $C^i$ using Equation \ref{equ:prob2}\\
   \eIf{ $\textbf{A}(\Delta,i) > \nu \ \forall i$}{
    Compute $\gamma_j$ using Equation \ref{equ:epsilon2} with $\textbf{b} = 1$\\
    }{
    Compute $\gamma_j$ using Equation \ref{equ:epsilon2} with $\textbf{b} = 2$\\
    }
   }
  Compute $\mu_{\gamma_j}(C^i|\widetilde{S}_{\tau_j})$ using Equation \ref{equ:assign2} \\
  $\forall C^i \in \textbf{C} = \{C^1,C^2,\ldots,C^K\}$ and $C^{K+1}$ \\
  Apply `stickiness' adjustment using Equation \ref{equ:dec} \\
  Compute $\Pr(C^{i}|\widetilde{S}_{\tau_j}) \ \forall i \in \{1,2,\ldots,C^{K+1}\}$ using Equation \ref{equ:norm} \\
  Assign $\widetilde{S}_{\tau_j}$ to a class via sampling from the distribution $\Pr(C^{i}|\widetilde{S}_{\tau_j})$ \\
  \uIf{$j\in \{1,2,\ldots,K\}$}{
  Update $N^j_{mn}$ by appending $\widetilde{S}_{\tau_l}$ to $S^j$\\
  }
  \ElseIf{$j = K+1$}{
  Update $\textbf{C}$ as $\{C^1,C^2,\ldots,C^K,C^{K+1}\}$\\
  Compute $N^{K+1}_{mn}$ using $\widetilde{S}_{\tau_l}$ \\
  }
  }
  \caption{Online $HSDF$ algorithm.}
\end{algorithm}

\vspace{10pt}
\subsection{Off--line PFSA revision} \label{sec:PFSA}
Algorithm \ref{algo:1} operates at the lowest logical layer in an online manner for learning multiple PFSA models representing different unique quasi--stationary characteristics. Representation of these characteristics is performed by considering the changes in the data likelihood and its rate of change. However, when the data quality is low especially in term of signal-to-noise ratio (SNR), the online learning algorithm may generate many spurious classes. In such cases, redundant PFSA models may be pruned periodically, that is, after a few slow time epochs have been observed. The pruning step proposed here merges different $PFSA$ models whose proximity are evaluated with the metric laid out in the Definition ~\ref{def:dis_metric} below (according to~\cite{MR14}).
\begin{definition}(Distance Metric for $PFSA$) \label{def:dis_metric}
Let $\mathcal{P}^1 = (\Theta^1, \Xi, \delta^1, \Omega^1)$ and $\mathcal{P}^2 = (\Theta^2, \Xi, \delta^2, \Omega^2)$ be two $PFSA$ with a common alphabet $\Xi$. Let $Pr_1(\Xi_r)$ and $Pr_2(\Xi_r)$ be the steady state probability vectors of generating words of length $r$ from the $PFSA$, $\mathcal{P}^1$ and $\mathcal{P}^2$, respectively, i.e., $Pr_1(\Xi_r) \triangleq [Pr(w)]_{w \in \Xi_r}$ for $\mathcal{P}^1$ and $Pr_2(\Xi_r) \triangleq [Pr(w)]_{w \in \Xi_r}$ for $\mathcal{P}^2$. Then, the metric for the distance between the $PFSA$ models, $\mathcal{P}^1$ and $\mathcal{P}^2$ is defined as
\begin{equation}
\Phi(\mathcal{P}^1, \mathcal{P}^2) \triangleq \lim_{n \to \inf} \sum_{r=1}^n \frac{\| Pr_1(\Xi_r) - Pr_2(\Xi_r) \|_{l_1}}{2^{r+1}}
\end{equation}
where the norm $\| \star \|_{l_1}$ indicates the sum of absolute values of the elements in the vector $\star$.
\end{definition}  
Thus, the pruning step can merge two different $PFSA$ models identified by online $HSDF$, $\mathcal{P}^1$ and $\mathcal{P}^2$ when $\Phi(\mathcal{P}^1, \mathcal{P}^2) < \eta$, where $\eta > 0$. In this paper, the metrics have been evaluated on symbols whose word length are $1$. Note that this revision step can be considered to be part of an off--line process for learning the higher-level (Tier 2) PFSA.
\nomenclature{$N$}{Symbol count matrix with training data}%
\nomenclature{$\widetilde{N}$}{Symbol count matrix with testing data}%
\nomenclature{$\mathbb{C}(\cdot)$}{CRP concentration function}%
\nomenclature{$\eta$}{Threshold for PFSA revision}%
\section{Improvement of Data Likelihood}\label{sec:Proof}
The algorithm proposed in this paper inherently aims to maximize the data likelihood as the $CRP$ formulation uses the likelihood of all the existing classes as concentration or strength function at any given epoch. When the likelihood values of the existing classes drop significantly, a new class is created to keep the data likelihood high with respect to the overall hierarchical model. Likelihood visualization in Section \ref{sec:validation} supports this notion as well. We observe that this process is equivalent to minimizing the Kullback-Liebler ($KL$) Divergence ~\cite{SR51}. Similar observations were made by~\cite{RB08,YB09}. Therefore, $KL$ Divergence $\rightarrow 0$ can be a relevant objective to learn the proposed hierarchical model and hence can be used for assuring that the algorithm can converge. Also, note that the stickiness adjustment and the $PFSA$ revision step aims to reduce the number of $PFSA$ at the lower layer without significant loss in data likelihood. Hence, the overall algorithm aims to minimize model complexity while capturing data characteristics (likelihood).\\
Before demonstrating the equivalence between data likelihood and $KL$ Divergence stated above, we present some mathematical preliminaries.
\subsection{Preliminaries}\label{sec:Prelims}
The gamma function $\Gamma(\alpha)$ can be expressed as,
\begin{align}\label{equ:gamma}
 \Gamma(\alpha) \triangleq e^{-\alpha}\alpha^{\alpha - \frac{1}{2}}\sqrt{2\pi}(1+\frac{1}{12 \alpha}+ \mathcal{O}(\frac{1}{\alpha^2}))
 \end{align}  
Using Stirling's approximation, the expression can be simplified under the assumption of $\alpha\gg\frac{1}{2}$ as the following:
\begin{align}\label{equ:apgam}
 \Gamma(\alpha) \approx e^{-\alpha}\alpha^\alpha
\end{align}  
Using this formula, we can rewrite the normalizing constant described in our online classification approach (see Equation \ref{equ:const}) as
\begin{align}\label{equ:beta}
 B( \boldsymbol{\alpha}^i_m )\approx\frac{\prod_{n=1}^{|\Xi|} e^{-\alpha^i_{mn}} (\alpha^i_{mn})^{\alpha^i_{mn}}}{e^{-(\sum_{n=1}^{|\Xi|} \alpha^i_{mn})} (\sum_{n=1}^{|\Xi|} \alpha^i_{mn})^{(\sum_{n=1}^{|\Xi|} \alpha^i_{mn})}}
\end{align}  
By eliminating common terms from the numerator and the denominator, we obtain
\begin{align}\label{equ:beta2}
 B( \boldsymbol{\alpha}^i_m )\approx\frac{\prod_{n=1}^{|\Xi|} (\alpha^i_{mn})^{\alpha^i_{mn}}} {(\sum_{n=1}^{|\Xi|} \alpha^i_{mn})^{(\sum_{n=1}^{|\Xi|} \alpha^i_{mn})}}
\end{align}  
Let the constant denominator term $(\sum_{n=1}^{|\Xi|} \alpha^i_{mn})^{(\sum_{n=1}^{|\Xi|} \alpha^i_{mn})}$ be denoted as $Z = ( N^i_{m} + |\Xi|)^{N^i_{m}+|\Xi|}$. With this setup Equation \ref{equ:fi} can be rewritten as
\begin{align} \label{equ:ni}
f_{\Omega^i|S^i}(\boldsymbol{\omega}^i|S^i)\approx \prod_{m=1}^{|\Theta|}Z \prod_{n=1}^{|\Xi|} \frac{(\boldsymbol{\omega}_m^i)^{N^i_{mn}}}{(N^i_{mn}+1)^{N^i_{mn}+1}}
\end{align}  
Similarly, at the testing stage, Equation \ref{equ:notation} can be rewritten as
\begin{align}\label{equ:note}
\Pr\left(\widetilde{S}|\Omega^i \right)\approx \prod_{m=1}^{|\Theta|}\widetilde{Z} \prod_{n=1}^{|\Xi|} \frac{(\Omega^i_{mn})^{\widetilde{N}_{mn}}}{(\widetilde{N}_{mn}+1)^{\widetilde{N}_{mn}+1}}
\end{align}   
where $\widetilde{Z} = (\widetilde{N}_{m})^{\widetilde{N}_{m}}$.
\begin{theorem} At a testing epoch, maximizing the loglikelihood $\log\Pr\left(\widetilde{S}|\Omega^i \right)$ is equivalent to minimizing $KL$ Divergence between the testing data distribution and training data distribution.
\end{theorem}
\textbf{Proof Sketch:}\label{sec:Prove}
\begin{align}
\log\Pr\left(\widetilde{S}|\Omega^i \right) \nonumber\\
&= \sum_{m=1}^{|\Theta|}\left(\log\widetilde{Z}-\sum_{n=1}^{|\Xi|}\left(\left[\widetilde{N}_{mn} +1\right] \log\left[\widetilde{N}_{mn} +1\right]- \widetilde{N}_{mn}\log\Omega_{mn}^i\right)\right)
\label{equ:prob6}
\end{align}  
After some algebraic rearrangement we obtain, 
\begin{align}
\log\Pr\left(\widetilde{S}|\Omega^i \right) = \sum_{m=1}^{|\Theta|}\left(-\sum_{n=1}^{|\Xi|}\left(\left[\widetilde{N}_{mn} +1\right]\log\left[\frac{\widetilde{N}_{mn} +1}{\Omega_{mn}^i}\right]\right)\right)  \nonumber\\
+ \sum_{m=1}^{|\Theta|} \left(\log(\widetilde{Z}) - \sum_{n=1}^{|\Xi|}  \log\Omega_{mn}^i\right) 
\label{equ:prob7}
\end{align}  
Note that $\sum_{m=1}^{|\Theta|}\log(\widetilde{Z})$ is a constant normalization factor term and $\Omega_{mn}^i$ represents the models learnt at the training stage for the existing classes (hence, does not change significantly). Therefore, if we aim to maximize the log-likelihood $\log\Pr\left(\widetilde{S}|\Omega^i \right)$ over all models (or classes) denoted by the index $i$, we obtain
\begin{align}
\arg\max_{i}\log\Pr\left(\widetilde{S}|\Omega^i \right) &\approx \arg\max_{i}\sum_{m=1}^{|\Theta|}\left(-\sum_{n=1}^{|\Xi|}\left(\left[\widetilde{N}_{mn} +1\right]\log\left[\frac{\widetilde{N}_{mn} +1}{\Omega_{mn}^i}\right]\right)\right)
\nonumber\\
&= \arg\max_{i}\sum_{m=1}^{|\Theta|}\left(-\sum_{n=1}^{|\Xi|}\left(\left[\widetilde{\alpha}_{mn}\right]\log\left[\frac{\widetilde{\alpha}_{mn}}{\Omega_{mn}^i}\right]\right)\right)  \nonumber\\
&= \arg\min_{i}\left(\text{$KL$  Divergence}\left(\widetilde{\alpha},\Omega^i\right)\right)
\label{equ:prob8}
\end{align} 

where $\widetilde{\alpha}$ represents the distribution of the testing data and $\Omega^i$ represents the training data distribution (for model $i$).
\section{Validation results and Discussion}\label{sec:validation}
The proposed algorithm is tested and validated in this section using data from simulated switching nonlinear dynamical systems. We begin with describing the simulation system and data generation scheme for validation.
\subsection{Simulated nonlinear dynamical systems}\label{sec:duffing}
\begin{figure} 
\centering
\includegraphics[width=\textwidth]{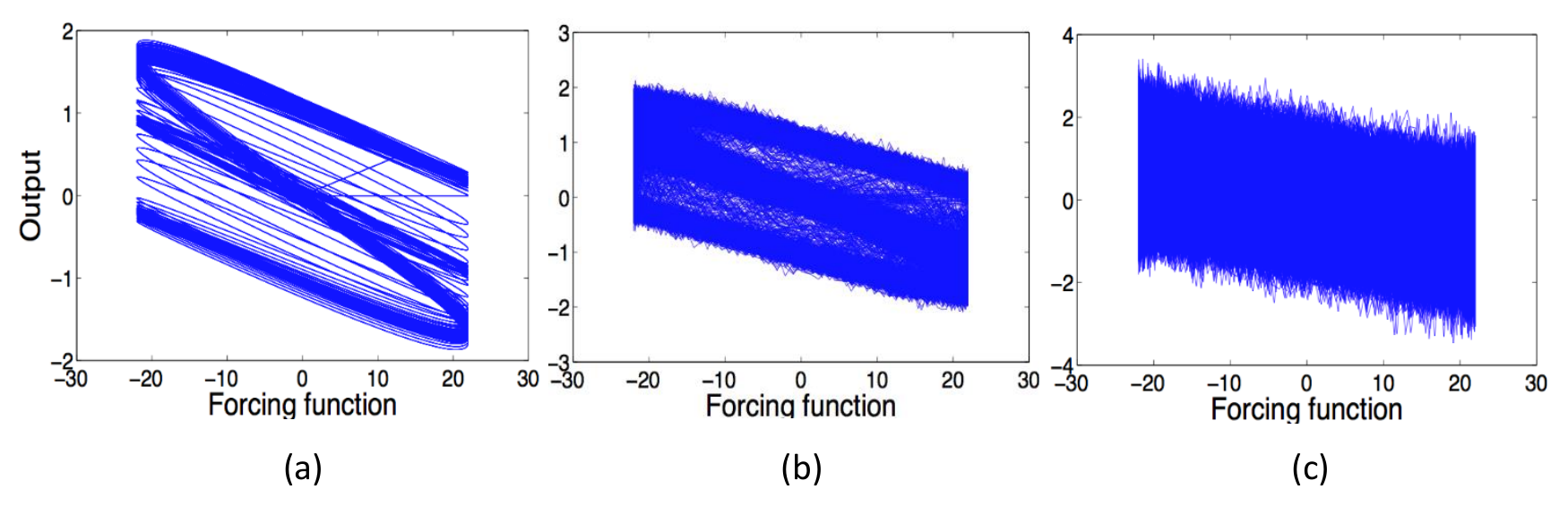}
\caption{\textit{Input-output plots of non-stationary dynamics for chaotic Duffing System under various signal to noise ratio ($SNR$) - Plate (a) $SNR = \infty$, Plate (b) $SNR = 9$ and Plate (c) $SNR = 1$.}}
\label{fig:Phaseplot}
\end{figure}

We use the chaotic Duffing system described by Equation \ref{equ:fm} which is a popular choice as a nonlinear system~\cite{RMSR09}. 
\begin{align}\label{equ:fm}
 \frac{d^2{x(t)}}{dt^2} + \beta \frac{dx(t)}{dt} + \alpha_1x(t) + \lambda x^3(t) = Acos(\textbf{w} t)
\end{align}
where $A = 22.0$ is the input amplitude, $\textbf{w}$ = 5.0 rad/s is its frequency of excitation, excitation harmonics, $\alpha_1$ = 1.0, stiffness, $\lambda$ = 1.0. It is know that varying $\beta$, the dissipation parameter causes change in the system behavior and a sudden shift or bifurcation occurs around $\beta = 0.3$ ~\cite{RRSY09}. Hence, $\beta = 0.1$ signifies an operating region before bifurcation and $\beta = 0.4$ represents a system behavior after bifurcation. Therefore, a non-stationary time series with two types of quasi-stationary segments can be generated by randomly selecting between the two $\beta$ values for different segments. Plots of the output $x$ vs. the forcing function are shown in Figure \ref{fig:Phaseplot} under different noise contamination levels.

\begin{figure} 
\centering
\includegraphics[width=\textwidth]{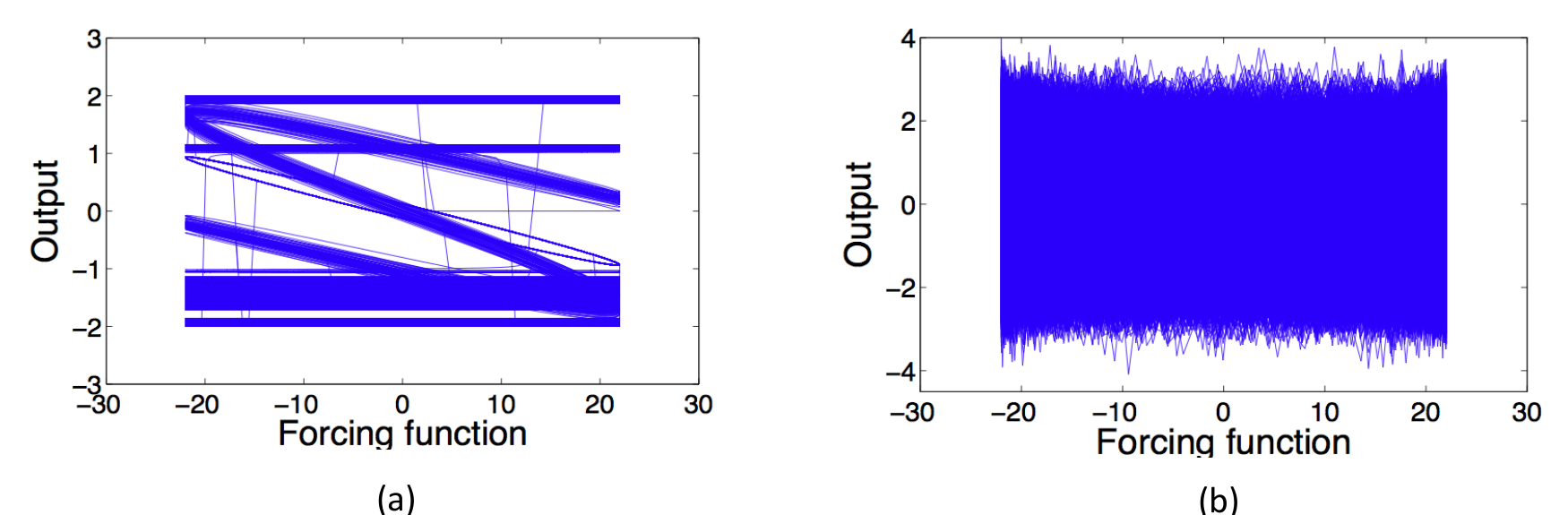}
\caption{\textit{Input-output plots of non-stationary dynamics for a random mix of chaotic Duffing system and Van der Pol system under various signal to noise ratio ($SNR$) - Plate (a) $SNR$ = $\infty$, Plate (b) $SNR$ = 1.}}
\label{fig:3phaseplot}
\end{figure}

We also extended the system to generate data with three types of quasi-stationary segments or features by adding a Van der Pol oscillation system behavior which is given by~\cite{MT06}
\begin{align}\label{equ:vanderpol}
 \frac{d^2{x(t)}}{dt^2} +1000 x^2(t) \frac{dx(t)}{dt} + x(t) = 1000
\end{align}
Figure \ref{fig:3phaseplot} shows the plots of the output $x$ vs. the forcing function with all the three features (two from the Duffing system and one from the Van der Pol system) under two different noise contamination levels. For both 2--features and 3--features cases, time series data with randomly generated $400$ epochs (with $1000$ data points of one particular feature in each epoch) are used for testing. Also for symbolic dynamic analysis, the raw time series is symbolized with a uniform partitioning (i.e., equal width binning) into $7$ bins that is found to be sufficient experimentally for most cases.

\subsection{Results and Performance comparison}\label{sec:results}
We evaluate the performance of the proposed algorithm on both the two and three features test cases. However, we begin with analysis to explore the effects of adaptive $CRP$ formulation with parameter $\textbf{b}$, stickiness adjustment and off--line revision steps.

\begin{figure} 
\centering
\includegraphics[width=\textwidth]{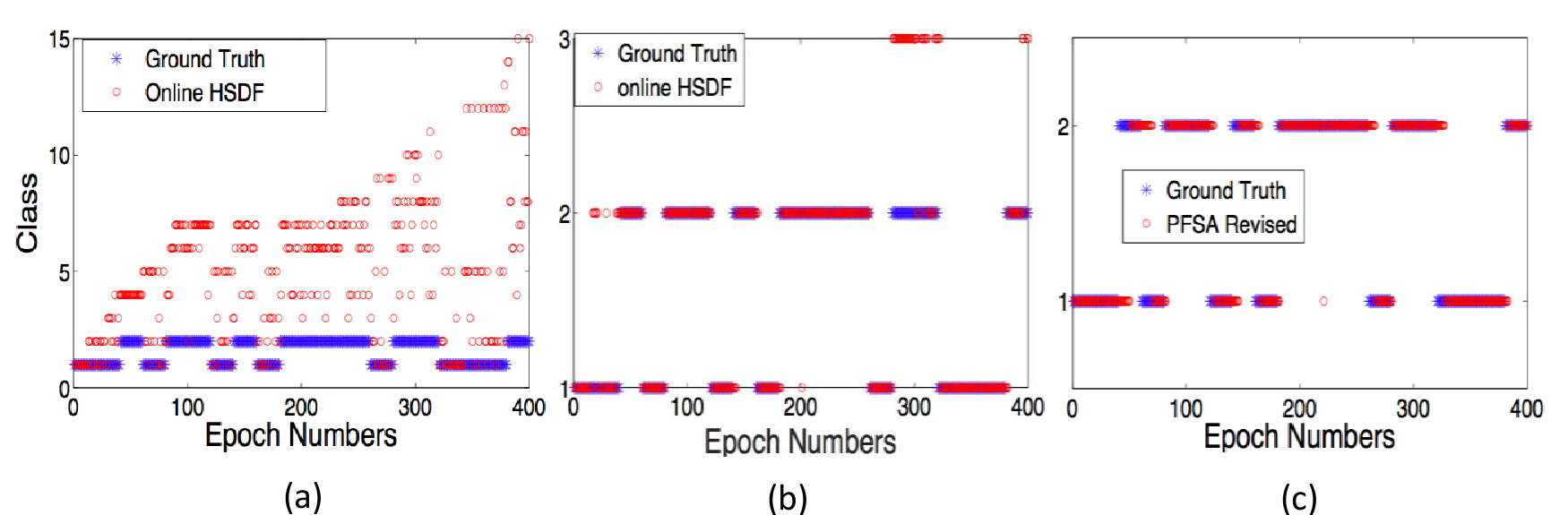}
\caption{\textit{Results for noiseless Duffing system where, Plate (a) $HSDF$ using classical $CRP$--only, Plate (b) $HSDF$ using classical $CRP$ and stickiness, and Plate (c) off--line $PFSA$ revision.}}
\label{fig:noiseless}
\end{figure}

\begin{figure} 
\centering
\includegraphics[width=\textwidth]{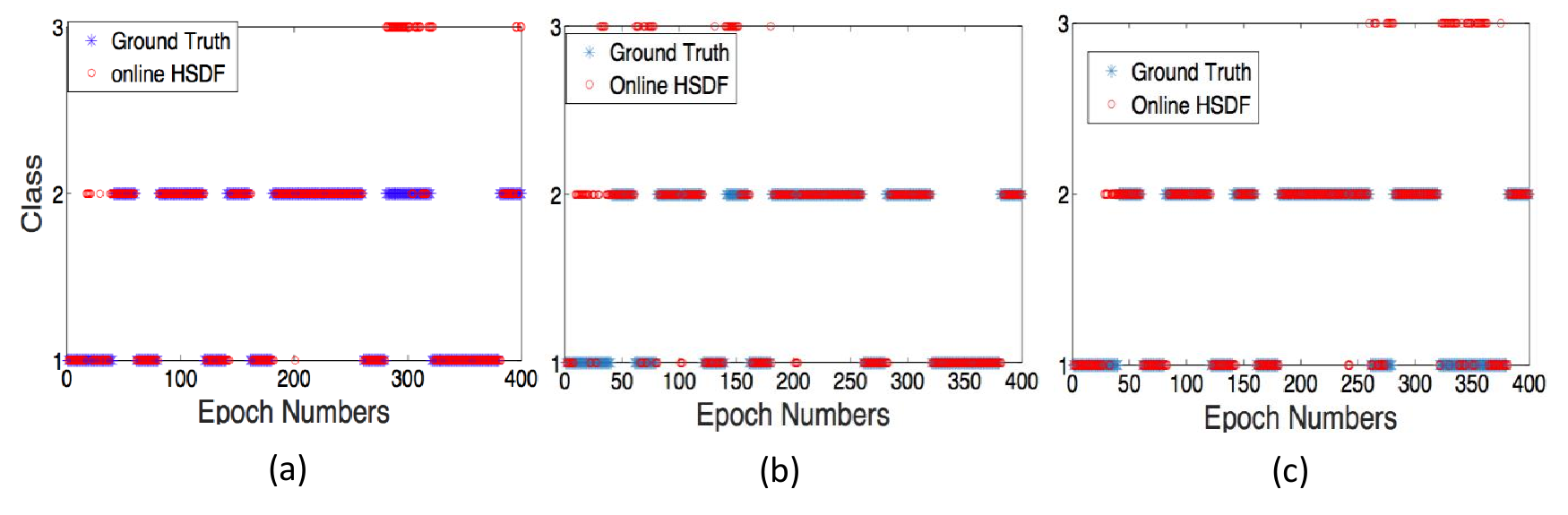}
\caption{\textit{Results for Duffing system using on-line $HSDF$ (i.e., without off-line $PFSA$ revision) with classical $CRP$ formulation under different noise contamination levels - Plate (a) $SNR = \infty$, Plate (b) $SNR = 9$ and Plate (c) $SNR = 1$.}}
\label{fig:classical}
\end{figure}

\textbf{Performance with classical $CRP$ formulation:} Many traditional methods only consider data likelihood for feature change detection or outlier detection purposes. However, in our proposed algorithm, we also utilize the change in data likelihood to decide on generating new models. We incorporate the effect of change in likelihood via using the adaptive $CRP$ formulation with parameter $\textbf{b}$. While in our algorithm, $\textbf{b}$ can take a value of $1$ or $2$ depending on the parameter $\textbf{A}(\Delta,i)$ ($\Delta$ is chosen to be $4$ for the results in this paper), the classical $CRP$ formulation would use a constant $\textbf{b} = 1$. Figure \ref{fig:noiseless} presents the results using the classical formulation, where plate (a) shows the performance only after applying the $CRP$ step, plate (b) shows the effect of stickiness adjustment and plate (c) provides the final result after the off-line revision step (using $\eta = \frac{1}{2K}$ given $K$ classes from the online part, i.e., after stickiness adjustment). While the $CRP$ step enables the framework to detect changes in time series characteristics, evidently, the stickiness adjustment is critical to control the `hunting' behavior and create `too many' new classes. Finally, the off--line $PFSA$ revision step helps to improve the result even further. Note, this result is obtained using a noiseless (i.e., signal to noise ratio, $SNR = \infty$) data set with two features. Figure \ref{fig:classical} presents the effect of noise content (for $SNR = \infty$, $SNR = 9$ and $SNR = 1$) on the performance using the same algorithm (i.e., constant $\textbf{b} = 1$). The results demonstrate visually that the algorithm is quite robust to significant noise contamination.

\begin{figure} 
\centering
\includegraphics[width=\textwidth]{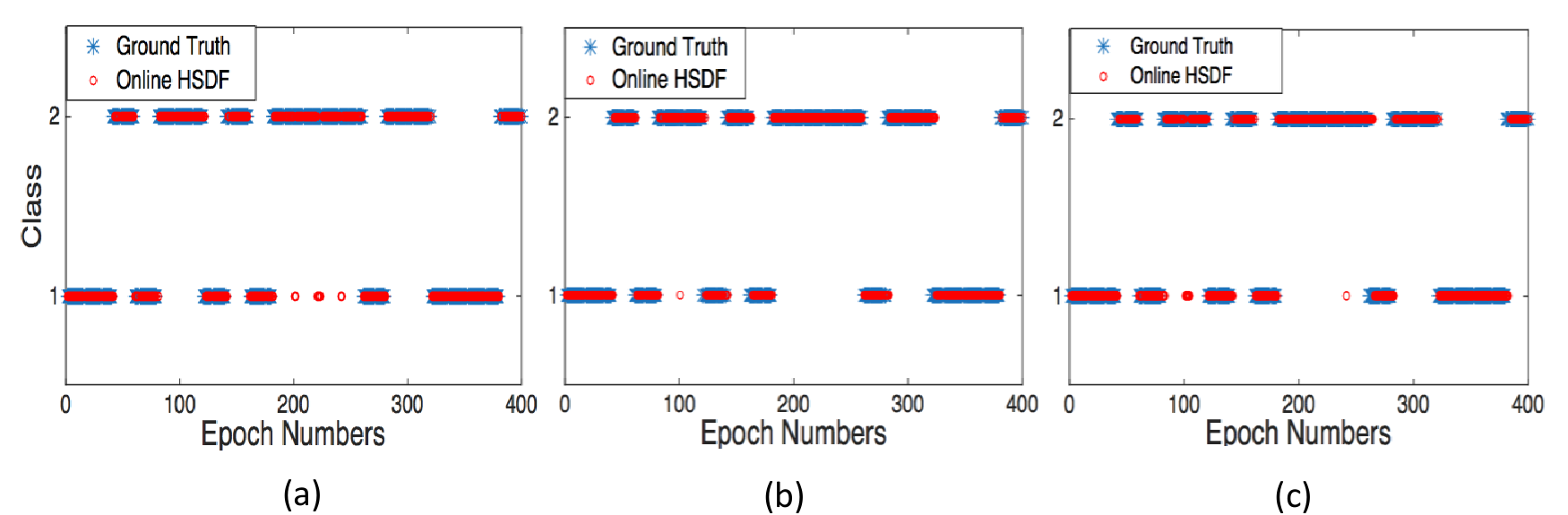}
\caption{\textit{Results for Duffing system using online $HSDF$ (i.e., without off-line PFSA revision) with adaptive $CRP$ formulation under different noise contamination levels - Plate (a) $SNR = \infty$, Plate (a) $SNR = 9$ and Plate (a) $SNR = 1$.}}
\label{fig:Improved_HSDF}
\end{figure}

\textbf{Performance with adaptive $CRP$ formulation:} Now we move to the adaptive $CRP$ formulation as described in Algorithm ~\ref{algo:1}, with an appropriate choice of $\textbf{b}$ (i.e., equal to 1 or 2) based on the change in data likelihood of the existing classes. Results are presented in Figure \ref{fig:Improved_HSDF} that show the adaptive formulation (with stickiness adjustment) to be quite efficient and achieves online performance better than that obtained after off--line revision with the classical formulation. Similar to the previous case, the algorithm is also quite stable under noise contamination. 

\begin{figure} 
\centering
\includegraphics[width=\textwidth]{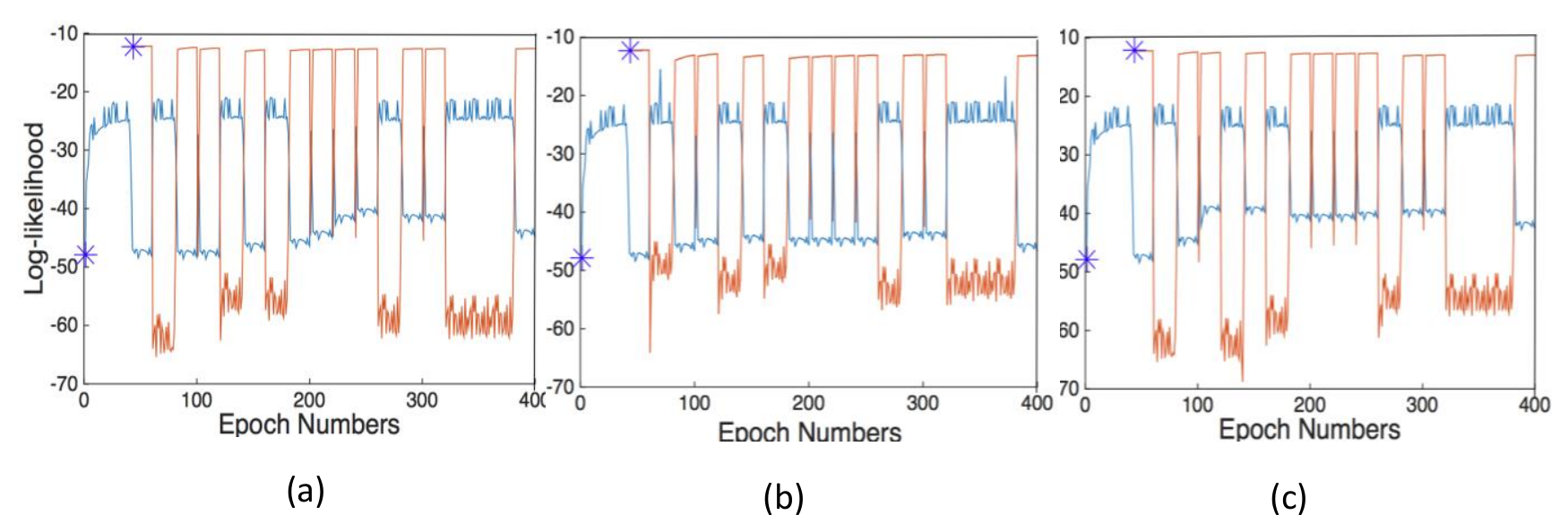}
\caption{\textit{Data log-likelihood plots with $\textcolor[rgb]{0.00,0.00,1.00}{\ast}$ representing the start of a new class for Duffing system using online $HSDF$ (i.e., without off-line $PFSA$ revision) with adaptive $CRP$ formulation under different noise contamination levels - Plate (a) $SNR = \infty$, Plate (a) $SNR = 9$ and Plate (a) $SNR = 1$.}}
\label{fig:Loglik}
\end{figure}

Additionally, we found that the sensitivity of the algorithm to the hyper-parameters, $\epsilon$ and $\kappa$ reduces significantly under the adaptive $CRP$ formulation. Typically, the values for $\epsilon$ and $\kappa$ used in this paper are $\approx 0.02$ and $\approx 0.6$ respectively. Figure \ref{fig:Loglik} shows the data log-likelihood plots for the class transitions and new class creation in an explicit manner. Table \ref{tab:sensitivity} compiles all the quantitative results for both classical and adaptive $CRP$ formulation under the different noise conditions considered here. The results show that online $HSDF$ with adaptive $CRP$ performs the best under low to moderate noise level. At a higher noise level, the off--line revision may be more suitable.
\begin{table}[!htp]
\centering
 \caption{Performance comparison of algorithm versions under different noise levels.}\label{tab:error} \vspace{3pt}
 \vspace{0pt}
\begin{tabular}{|c|c|c|c|}
  \hline
  &\multicolumn{3}{|c|}{\textbf{Error \%}}\\
  \hline
  \textbf{Algorithm}& $SNR = \infty$ & $SNR = 9$ & $SNR = 1$\\
   \hline
  $HSDF$ + Classical $CRP$& 12.94& 16.50& 16.50\\
   \hline
  $HSDF$ + Classical $CRP$ + $PFSA$ revision& 5.50& 6.25& \textbf{6.25}\\
  \hline
  $HSDF$ + Adaptive $CRP$& \textbf{4.75} & \textbf{5.75} & 7.00\\ 
   \hline
\end{tabular}
\label{tab:sensitivity}
\end{table}

\begin{figure} 
\centering
\includegraphics[width=\textwidth]{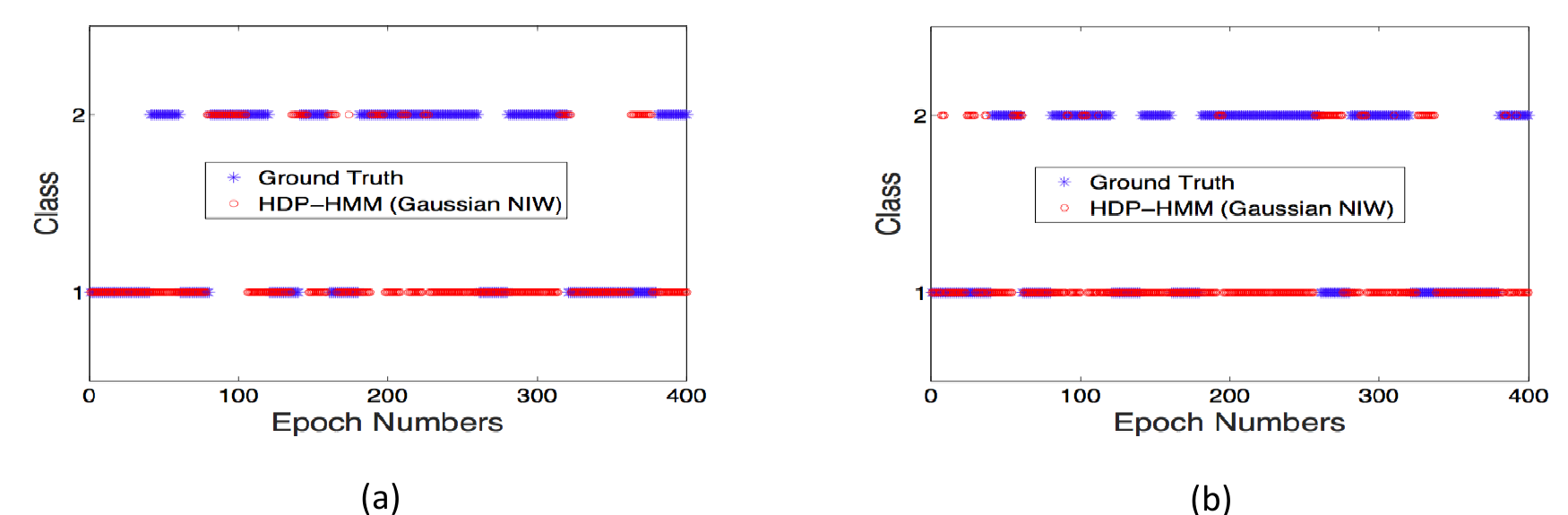}
\caption{\textit{Performance of HDP--HMM approach for time series with two features under different noise contamination levels - Plate (a) $SNR = \infty$, and Plate (b) $SNR = 1$.}}
\label{fig:duffing3_1}
\end{figure}

\textbf{Performance comparison:} We compared the results from our proposed algorithm with those from Hierarchical Dirichlet Process -- Hidden Markov Model (HDP--HMM) ~\cite{FSJW11} that is a sampling based technique based on the Bayesian nonparametric concept, such that the joint distribution of the states are derived from the Dirichlet process. HDP--HMM techniques have been used in learning switching linear dynamical systems ($SLDS$). Note that the idea of stickiness has been adopted from the HDP--HMM literature as we aim to extract features for more realistic and general cases of nonlinear dynamical systems in a fast and computationally efficient manner.
For comparison purposes, we implemented the codes made available on the authors' webpage~\footnote{https://www.stat.washington.edu/~ebfox/software.html} that uses a Gaussian observed model type with Normal--Inverse--Wishart ($NIW$) prior, which we found to be producing best results for all use cases. The results for the HDP--HMM approach under the noiseless ($SNR = \infty$) and the most noisy ($SNR = 1$) cases are shown in Figure \ref{fig:duffing3_1}. Note that the HDP--HMM \cite{FSJW11} algorithm classifies each data point individually, being a sampling technique. Hence, a majority voting was done to select the most prominent class in each epoch (i.e., $1000$ data points as defined earlier) for a more realistic comparison. From the plots, it is quite evident visually that our proposed approach has better accuracy which can be explained by the fact that $SDF$ inherently is an efficient way to model nonlinear system behavior. 

\begin{figure} 
\centering
\includegraphics[width=\textwidth]{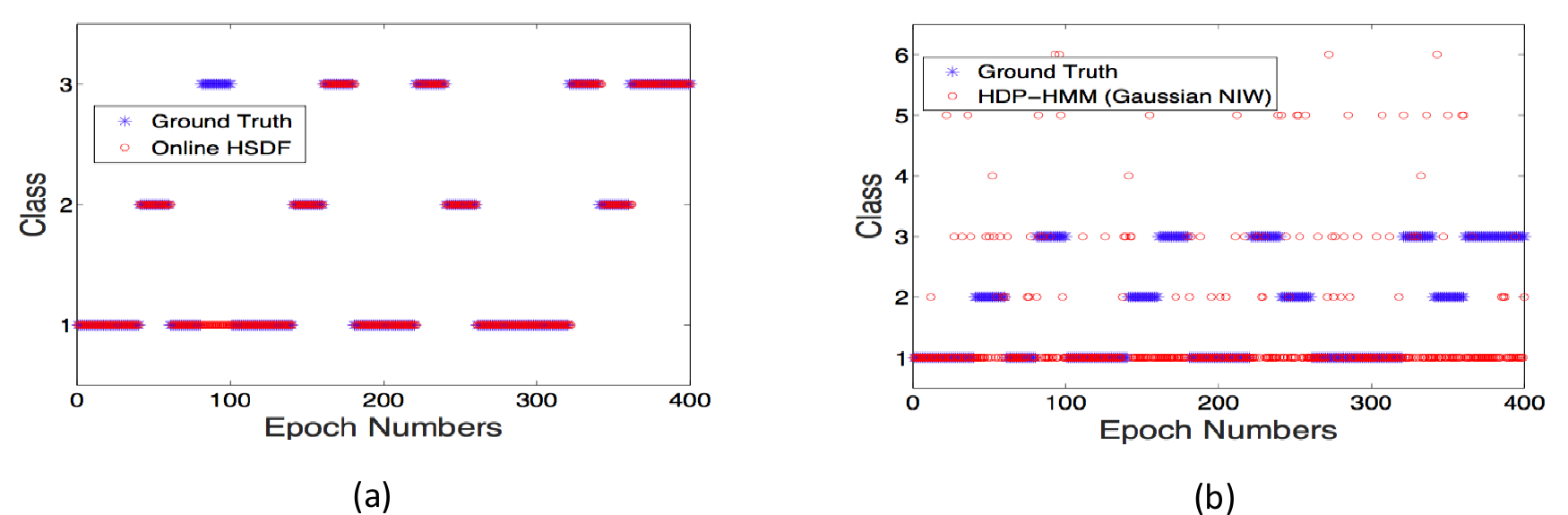}
\caption{\textit{Performance comparison of (a) $HSDF$ and (b) HDP--HMM approaches for timeseries with three features under no noise condition.}}
\label{fig:duffing3_2}
\end{figure}

\begin{figure}[b] 
\centering
\includegraphics[width=\textwidth]{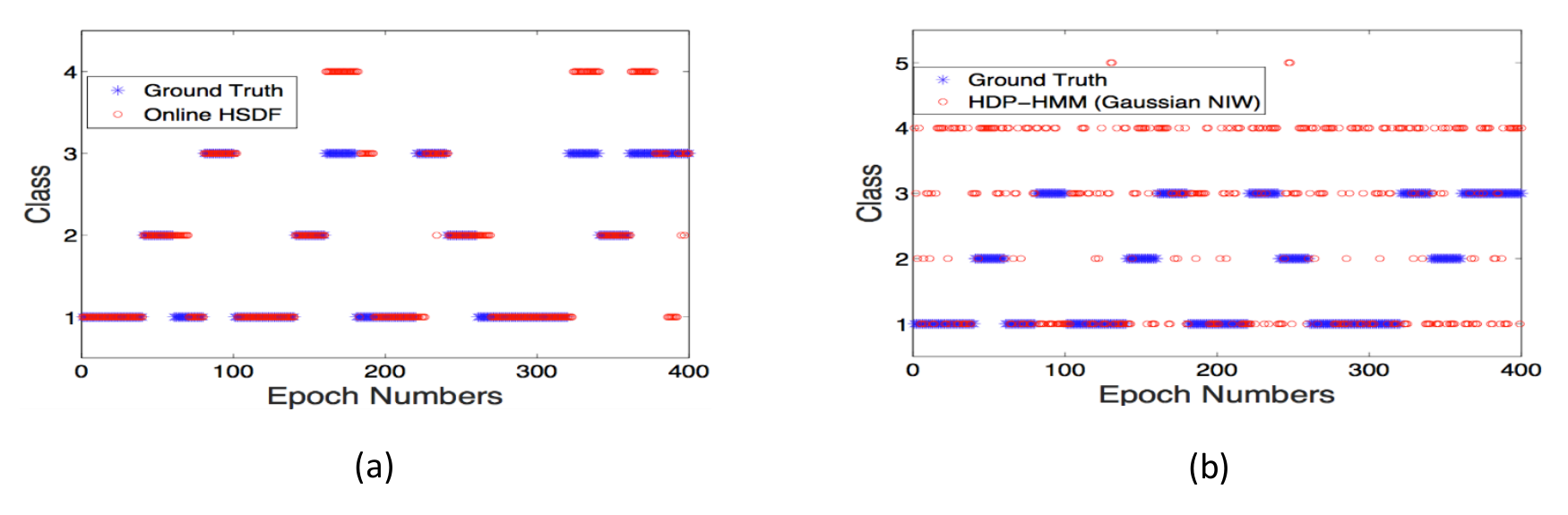}
\caption{\textit{Performance comparison of (a) $HSDF$ and (b) HDP--HMM approaches for time series with three features under $SNR = 1$ condition.}}
\label{fig:duffing3_3}
\end{figure}

Also, the performance of the HDP-HMM approach suffers significantly in the presence of a large amount of noise in the signal. However, we note that HDP--HMM approach correctly identifies the number of features present in the non--stationary time series data which is significant. We performed further comparison using the 3--features case, and present the results in Figure \ref{fig:duffing3_2} and Figure \ref{fig:duffing3_3} for the $SNR = \infty$ and $SNR = 1$ cases respectively. It can be observed that our algorithm still performs better in both the identification of the total number of features present in the time series data as well as in classifying them. Finally, the quantitative performance of $HSDF$ and HDP--HMM (Gaussian $NIW$) for the two and three features cases are summarized in Table \ref{tab:performance}. Note, the smaller time requirement for our proposed algorithm compared to that of HDP--HMM is primarily due to the fact that there is no sampling step or latent variable involved in the HSDF approach. The computation time reported here is achieved with MATLAB implementations on a $2.1$ GHz Intel Xeon(R), $1200$ MHz CPU with $64$GB RAM and UNIX OS.

\begin{table}[h]
\centering
\caption{Performance comparison of $HSDF$ and HDP--HMM approaches.}
\label{tab:performance}\vspace{3pt}
\vspace{0pt}
\begin{center}
\begin{tabular}{|c|c|c|c|c|c|}
      \hline
      \multicolumn{2}{|c|}{\textbf{Method}}& \multicolumn{2}{|c|}{\textbf{Online HSDF}} &\multicolumn{2}{|c|}{\textbf{HDP--HMM (Gaussian NIW)}}\\
      \hline
      \multicolumn{2}{|c|}{\textbf{Noise level}}&$SNR$ = $\infty$ & $SNR$ = $1$&$SNR$ = $\infty$ & $SNR$ = $1$\\
      \hline
      \multirow{2}{*}{2 features}&Error($\%$)&$4.75$ &$7.00$ &$48.25$ &$57.50$ \\
      &Time($secs$)&$66.1$&$65.0$&$423.0$ &$419.7$ \\%
      \hline
      \multirow{2}{*}{3 features}&Error($\%$)&$7.25$ & $26.5$&$56.25$ &$76.75$\\
      &Time($secs$) &$88.1$ &$112.4$ &$418.0$ &$426.8$\\%
      \hline
\end{tabular}
\end{center}
\end{table}
\section{Summary, Conclusions and Future work}\label{sec:con}
This paper builds on the concepts of Symbolic Dynamic Filtering ($SDF$) of modeling quasi--stationary time series to frame a computationally simple, efficient technique for extracting hierarchical features from slow time--scale non--stationary time series data that comprises of quasi--stationary time series segments. We use the concepts of probabilistic finite state automata ($PFSA$), Chinese Restaurant Process ($CRP$), stickiness and likelihood change rate to create the proposed hierarchical framework with self--similar layers. While we capture multiple quasi--stationary dynamics at a fast time--scale using individual $PFSA$ models at a lower layer, transitions of the system at a slow time--scale among different quasi--stationary dynamics are captured using similar $PFSA$ model at a upper layer. Also note that the developed algorithm enables unsupervised data analysis where the number of unique quasi--stationary behaviors present in the data is unknown. Although the primary learning goal is to consistently improve the data likelihood with the overall hierarchical model (or reduce the Kullback--Leibler ($KL$) divergence between the model and the data distributions), we show that tracking the change in likelihoods of different unique quasi--stationary characteristics leads to a more efficient algorithm. We accommodate this new feature using a novel adaptive $CRP$ formulation. The proposed algorithm is tested and validated using time series data generated from well--known nonlinear dynamical system simulation involving Duffing and Van der Pol equations. We demonstrate the efficacy of the algorithm under various noise contamination levels and in comparison with the competing HDP--HMM approach. We note that a key advantage of the proposed technique is its low computational and memory complexity. Hence, it can be extremely suitable for on--board real time applications.

While we formulate the central algorithm for $HSDF$ learning in this paper, we just show a two-tier modeling scheme here. Therefore, a key next step is to show the learning of a multi-layer (with more than two layers) $HSDF$s which is currently being pursued. A few other future research topics are:
\begin{itemize}
\item \emph{Extension of $HSDF$ algorithm for multivariate time series data in a scalable manner}
\item \emph{Optimal learning of heterogeneous $PFSA$ (i.e., different $PFSA$ structures for different quasi-stationary characteristics) at a lower layer}
\item \emph{Comprehensive testing and validation on real data sets}
\end{itemize}
\section*{Acknowledgement}
This work has been supported in part by the National Science Foundation (NSF) under Grant number CNS--\underline{1464279}.

\bibliography{DeepSDF}

\end{document}